\documentstyle[11pt,psfig]{article} 

\input{epsf}

\begin{document}

\title{\large\bf
\vspace*{-.5in} Robust Combining of Disparate Classifiers 
through Order Statistics \\ [.1in] }
\author{ \\
{\bf Kagan Tumer} \\
NASA Ames Research Center \\
MS 269-2, Moffett Field, CA, 94035-1000 \\
kagan@ptolemy.arc.nasa.gov \\
{\bf Joydeep Ghosh} \\
Department of Electrical and Computer Engineering, \\
University of Texas,  Austin, TX 78712-1084 \\
ghosh@pine.ece.utexas.edu \\
}
\date{\today}
\maketitle

\thispagestyle{empty}

\abstract{
Integrating the outputs of multiple classifiers via combiners
or meta-learners has led to
substantial improvements in several difficult pattern recognition
problems. 
In the typical setting investigated till now, each classifier is trained
on data taken or resampled from
a common data set, or (almost) randomly selected subsets thereof, and
thus experiences similar quality of training data.
However, in certain situations where data is acquired and analyzed on-line at
several geographically distributed locations, the quality of data
may vary substantially, leading to large discrepancies in 
performance of individual classifiers. 
In this article we introduce and investigate 
a family of classifiers based on order statistics, for robust handling
of such cases.
Based on a mathematical modeling of how the decision boundaries
are affected by order statistic combiners, we derive
expressions for the reductions in error expected when such combiners
are used.
We show analytically that the selection of the median, the maximum
and in general, the $i^{th}$ order statistic improves classification
performance.  
Furthermore, we introduce the trim and spread combiners, both
based on linear combinations of the ordered classifier outputs, and
show that they are quite 
beneficial in presence of outliers or uneven classifier performance.
Experimental results on several public domain data sets
corroborate these findings.  }

\section{Introduction}
\label{sec:intro}
Since different types of classifiers have different ``inductive bias'',
one does not expect the generalization performance of two classifiers to be
identical~\cite{gebi92,ghtu94a} for difficult pattern recognition
problems, even when they are both 
trained on the same data set. If only 
the ``best'' classifier is selected based on an
{\em estimation} of the true generalization performance 
using a finite test set~\cite{wolp90b}, valuable 
information contained in the results of the discarded 
classifiers may be lost.
Such potential loss of information can be avoided if 
the outputs of all available classifiers are used
in the final classification decision.
This concept has received a great deal of attention recently,
and many methods for combining  classifier outputs have been proposed
~\cite{ghbe92,hasa90,hohu94,peco93,shar96a}.
Furthermore, diversity among classifiers has been actively promoted,
by strategies such as bagging~\cite{brei94},
arcing~\cite{brei96,frsc95,frsc96}, 
boosting~\cite{drco94,drco96,quin96,scha90,scfr97}, 
and correlation control~\cite{alpa95a,tugh96b},
as a prelude to combining.

Approaches to pooling classifiers can be separated
into two main categories: (i) simple combiners, e.g., 
voting~\cite{baco94,chst95}, Bayesian based weighted 
product rule~\cite{kiha98}, 
or averaging~\cite{peco93b,tugh96}, 
and, (ii) meta-learners, such as arbitration~\cite{chst97} or
stacking~\cite{brei93,wolp92}. 
The simple combining methods are best suited for problems
where the individual classifiers perform the same task, and have
comparable success. However, such combiners are more susceptible to outliers
and to unevenly performing classifiers.
In the second category, 
either sets of combining rules, or
full fledged classifiers acting on the outputs of
the individual classifiers,  are constructed~\cite{alku95,jaco95,wolp92}.
This type of
combining is more general, but is vulnerable to all the problems associated
with the added learning (e.g., overparameterizing, lengthy training time).

An  implicit assumption in most combining schemes is that each 
classifier sees the same training data or resampled versions of the same
data. If the individual classifiers are then appropriately chosen
and trained properly, their performances will be (relatively) comparable
in any region of the problem space.
So gains from combining are 
derived from the diversity~\cite{krve95,opsh96} among classifiers 
rather that by compensating for weak members of the pool.
However, in real life, there are situations where individual classifiers 
may not have access to the same data. Such conditions arise
in certain data mining, sensor fusion and electrical logging (oil services)
problems where 
there are large variabilities in the data which is
acquired locally and needs to be processed in (near) real time 
at geographically separated places \cite{dasa94}. 
These conditions
create a pool of 
classifiers that may have significant variations in their overall
performance.
Moreover, they may lead to conditions where
individual classifiers have similar {\em average} performance,
but substantially different performance over different parts
of the input space. 

In such cases, combining is still desirable, but neither
simple combiners nor meta-learners are particularly well-suited for
the type of problems that arise.
For example, the simplicity of averaging the classifier outputs is appealing,
but the prospect of one poor classifier corrupting the combiner
makes this a risky choice.
Weighted averaging of classifier outputs appears to provide some 
flexibility~\cite{hasc93,mepa97}. 
Unfortunately, the 
weights are still assigned on a per classifier 
basis 
rather than a per sample or per class basis.
If a classifier is accurate only in
certain areas of the input space, this scheme fails to take
advantage of the variable accuracy of the classifier in question.
Using a meta learner that provides 
different weights for different patterns can potentially solve this problem,
but at a considerable cost. In particular, the off-line training of a meta-learner
using substantial amount of data outputted by geographically distributed classifiers,
may not be feasible.
In addition to providing robustness, the order statistic 
combiners presented in this work also 
aim at bridging the gap between simplicity and generality
by allowing the flexible selection of classifiers 
without the associated cost of training meta-classifiers.

Section~\ref{sec:error} summarizes the relationship between classifier
errors and decision boundaries and provides the necessary background
for mathematically analyzing order statistic combiners~\cite{tugh96}. 
Section~\ref{sec:order} introduces simple order statistic
combiners.
Based on these concepts, in Section~\ref{sec:comb} we 
propose two powerful combiners, {\em trim} and {\em spread}, 
and derive the amount of  error reduction
associated with each.
In Section~\ref{sec:res} we present the performance of
order statistic combiners on Proben1/UCI benchmarks~\cite{prec94}.
Section~\ref{sec:disc} discusses the implications of using
linear combinations of order statistics as a strategy for
pooling the outputs of individual classifiers.

\section{Error Characterization in a Single Classifier}
\label{sec:error}
In this section we summarize the approach and results of 
\cite{tugh96}\footnote{This and other related papers can be downloaded from
URL {\em http://www.lans.ece.utexas.edu}.},
that quantify the effect of inaccuracies in estimating {\em a posterior}
class probabilities on the classification error for a {\em single} classifier.
This background is needed to characterize and understand 
the impact of order statistics combiners, as described in Sections 3 and 4.

It is well known that, given {\em one-of-L} desired outputs and sufficient training samples
reflecting the class priors, the outputs of certain classifiers 
trained to minimize a mean square or cross-entropy error criteria, approximate the 
{\em a posteriori} probability densities of the corresponding 
classes~\cite{rili91,ruro90}.
Based on this result,
one can model the $i$th output of the $m$th such classifier as:
\begin{eqnarray}
f_i^{m}(x) = p_i(x) + \epsilon_i^m(x) ,
\label{eq:fmod}
\end{eqnarray}
where $p_i(x)$ is the  true posterior for $i$th class on input $x$, and
$\epsilon_i^m(x)$ is the error of the $m$th classifier in
estimating that posterior. 

\begin{figure}[htb]
\centerline{\psfig{figure=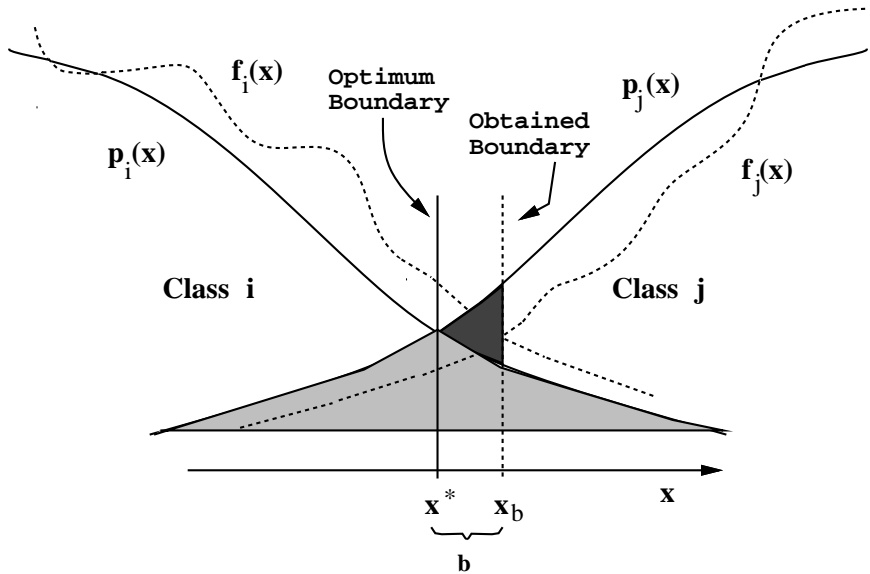}}
  \caption{Error regions associated with approximating the
      {\em a posteriori} probabilities~\protect\cite{tugh96}.}
  \label{fig:bd}
\end{figure}

Now, let us decompose the error into
two parts: $\epsilon_i^m(x) = \beta_i^m + \eta_i^m(x)$.
The first component does not vary with the input, and provides an
offset, or systematic error for each class. 
The second component gives the variability
from that systematic error, for each $x$ in each class, 
and has zero mean and variance $\sigma^2_{\eta_i^m(x)}$. 
These two components of the error are similar to the bias
and variance decomposition for a quadratic loss function
given in \cite{gebi92}, although they are
at the individual input level. We will therefore refer to
classifiers as ``biased'' and ``unbiased''  implying $\beta_k^m \neq 0$
for some $k,m$, and $\beta_k^m = 0 \; , \:  \forall k,m$, respectively.
Let $b^m$ denote the offset between the ideal class boundary, $x^*$ (based on
$p_i(x) = p_j(x)$) and the realized boundary, $x_b^m$ 
(based on $f_i^m(x) = f_j^m(x)$), as shown in 
Figure~\ref{fig:bd} \cite{tugh96}.  This boundary offset ($b^m = x_b^m - x^*$)
has mean and variance given respectively by: 
\begin{eqnarray}
\beta^m = \frac{\beta_i^m - \beta_j^m}{s} ,
\label{eq:meanb}
\end{eqnarray}
and
\begin{eqnarray}
\sigma^2_{b^m} = \frac{\sigma^2_{\eta_i^m(x)} + \sigma^2_{\eta_j^m(x)}}{s^2} , 
\label{eq:sigb}
\end{eqnarray}
where $ s \: = \: p^\prime_j(x^*) \: - \: p^\prime_i(x^*) $
as introduced in~\cite{tugh96}.

Let us further denote the probability density function of this
boundary offset by $f_b(x)$.
The expected model error associated with the
selection of a particular classifier $m$, can then be expressed as:
\begin{eqnarray}
E_{model}^m = \int_{-\infty}^\infty A(b) f_b(b) db,
\label{eq:Ab}
\end{eqnarray}
where $A(b) = \int_{x^*}^{x^*+b} \left( p_j(x) - p_i(x) \right)dx$
is the error due to the selection of a particular decision
boundary.
In general, it is not possible to obtain the density function
for the boundary offset without making assumptions on the distributions
of the errors. 
However, a first order approximation, derived in \cite{tugh96}, leads to: 
\begin{eqnarray}
E_{model}^m = \int_{-\infty}^\infty \frac{1}{2} b^2 s f_b(b) db .
\end{eqnarray}
Let us define the first and second moments of the boundary offset as follows:
\begin{eqnarray*}
{\cal M}_1 \: =\: \int_{-\infty}^\infty x f_b(x) dx
\; \; \; \; \;  and \; \; \; \; 
{\cal M}_2 \: =\: \int_{-\infty}^\infty x^2 f_b(x) dx.
\end{eqnarray*}
If the individual classifiers are unbiased, 
the offset $b^m$ of a single classifier 
has ${\cal M}_1 = 0$ and ${\cal M}_2 = \sigma^2_{b^m}$, leading to:
\begin{eqnarray}
E_{model}^m \: = \: 
\frac{s \: {\cal M}_2}{2} \: = \: \frac{s \sigma^2_{b^m}}{2}.
\label{eq:err}
\end{eqnarray}
Now, if the classifiers are biased,
the variance of $b$ is left unchanged (given by
Equation~\ref{eq:sigb}), but the mean becomes
$\beta = \frac{\beta_i - \beta_j}{s}$. In other words, we have
${\cal M}_1 = \beta^m$ and $\sigma^2_{b^m} = {\cal M}_2 - {{\cal M}_1}^2$, 
leading to the following model error:
\begin{eqnarray}
E_{model}^m (\beta) = \frac{s {\cal M}_2}{2}\; = \; 
\frac{s}{2} \: (\sigma^2_{b^m} + ({\beta^m})^2).
\label{eq:bising}
\end{eqnarray}
To emphasize the distinction between biased and unbiased classifiers, the 
model error will be given as a function of $\beta$ for biased classifiers.
A more detailed derivation
of class boundaries and error regions is presented in ~\cite{tugh96}.
For analyzing the error regions after combining and comparing
them to the single classifier case, one needs to determine how
the first and second moments of the boundary
distributions are affected by combining. 
The following sections focus on obtaining those values for various
combiners.

\section{Combining Multiple Classifiers through Order Statistics}
\label{sec:order}
\subsection{Basic Concepts}
In this section, we briefly discuss some basic concepts and
properties of order statistics. Let $X$ be a random variable
with probability density function $f_X(\cdot)$, and cumulative distribution
function $F_X(\cdot)$. Let $(X_1,X_2,\cdots,X_N)$ be a random sample drawn
from this distribution.
Now, let us arrange them in non-decreasing order, providing: 
\begin{eqnarray*}
X_{1:N} \leq  X_{2:N} \leq \; \cdots \; \leq X_{N:N}.
\end{eqnarray*}
The $i$th order statistic denoted by $X_{i:N}$, is the $i$th value
in this progression. The cumulative distribution function for the smallest
and largest order statistic can be obtained by noting that:
\begin{eqnarray*}
F_{X_{N:N}}(x) = P(X_{N:N} \leq x) = \Pi_{i=1}^N P(X_{i:N} \leq x) = [F_X(x)]^N \end{eqnarray*}
and:
\begin{eqnarray*}
F_{X_{1:N}}(x) & = &  P(X_{1:N} \leq x) = 1 - P(X_{1:N} \geq x) = 
1 - \Pi_{i=1}^N P(X_{i:N} \geq x) \\
& = & 1 - (1 - \Pi_{i=1}^N P(X_{i:N} \leq x) = 1 - [ 1 - F_X(x)]^N 
\end{eqnarray*}
The corresponding probability density functions can be obtained from
these equations. 
In general, for the $i$th order statistic, the cumulative distribution
function gives the probability that exactly $i$ of the chosen $X$'s are
less than or equal to $x$. 
The probability density function of $X_{i:N}$ is then
given by \cite{davi70}:
\begin{eqnarray}
f_{X_{i:N}}(x) = \frac{N!}{(i-1)! \: (N-i)!} \left[F_X(x) \right]^{i-1} 
\left[1 - F_X(x)\right]^{N-i} f_X(x) \; .    
\label{eq:osdensity}
\end{eqnarray}
This general form however, cannot always be computed in closed form. 
Therefore, obtaining
the expected value of a function of $x$ using Equation~\ref{eq:osdensity}
is not always possible. However, the first two moments of the density
function are widely available for a variety of distributions \cite{arba92}.
These moments can be used to compute the expected values of certain specific
functions, e.g., polynomials of order less than two.

\subsection{Combining Unbiased Classifiers through Order Statistics}
\label{sec:osno}
Now, let us turn our attention to order statistics (OS) combiners. 
For a given input $x$, let the network outputs of each of the $N$ 
classifiers for each class $i$ be ordered in the following manner:
\begin{eqnarray*}
f^{1:N}_i(x) \leq  f^{2:N}_i(x) \leq
\; \cdots \; \leq f^{N:N}_i(x).
\end{eqnarray*}
Then one constructs the $k$th order statistic combiner, by
selecting the $k$th ranked output for each class ($f^{k:N}_i(x)$), as 
representing its posterior~\cite{tugh95f}.

In particular, $max$, $med$ and $min$ combiners are defined 
as follows:
\begin{eqnarray}
\label{eq:max}
f_i^{max}(x)\; & = & \; f^{N:N}_i(x), \\
\label{eq:med}
f_i^{med}(x)\; & = & \; \left\{ \begin{array}{ll}
     \frac {f^{\frac{N}{2}:N}_i(x) \; + \:
f^{\frac{N}{2}+1:N}_i(x)}{2} & \mbox{if $N$ is even} \\
f^{\frac{N+1}{2}:N}_i(x)           & \mbox{if $N$ is odd},
                       \end{array} \right. \\
\label{eq:min}
f_i^{min}(x)\; & = & \; f^{1:N}_i(x). 
\end{eqnarray}
These three combiners are relevant because they represent important
qualitative interpretations of the output space. Selecting the maximum
combiner is equivalent to selecting the class with the highest
posterior. Indeed, since the network outputs approximate the
class {\em a posteriori} distributions, selecting the maximum 
reduces to selecting the classifier that is the most ``certain''
of its decision. The drawback of this method however is that it can
be compromised by a single classifier that repeatedly provides high
values. The selection of the minimum combiner follows a similar logic, but 
focuses on classes that are unlikely to be correct, rather than on the
correct class. Thus, this combiner eliminates less likely classes by basing 
the decision on the lowest value for a given class. This combiner suffers
from the same ills as the {\em max} combiner. However, it is 
less dependent on a single error, since it performs a min-max operation,
rather than a max-max\footnote{Recall that the pattern is ultimately
assigned to the class with the highest combined output.}.  
The median classifier on the other hand considers the most ``typical'' 
representation of each class. For highly
noisy data, this combiner is more desirable than either the {\em min}
or {\em max} combiners since the decision is not compromised as much
by a single large error.

The analysis that follows 
does not depend on the particular order statistic chosen. 
Therefore, we will denote all OS combiners by $f_k^{os}(x)$
and derive the model error, $E_{model}^{os}$.
The network output provided by
$f_k^{os}(x)$ is given by:
\begin{eqnarray}
f^{os}_k(x) = p_k(x) + \epsilon^{os}_k(x) \:,
\label{eq:os}
\end{eqnarray}

Let us first investigate the zero-bias case ($\beta_k = 0 \:, \: \forall k$), 
where we get $\epsilon^{os}_k(x) = \eta^{os}_k(x)$. 
Proceeding as in Section~\ref{sec:error}, the boundary $b^{os}$ is shown to be:
\begin{eqnarray}
b^{os} \: = \: \frac{\eta^{os}_i(x_b) - \eta^{os}_j(x_b)}{s}.
\label{eq:bos}
\end{eqnarray}
For {\em i.i.d.} $\eta_k$'s, the first two moments 
will be identical for each class. 
Moreover, taking the
order statistic will shift the mean of both $\eta^{os}_i$ and $\eta^{os}_j$
by the same amount, leaving the mean of the difference unaffected. 
Therefore, $b^{os}$ will have zero mean, and variance:
\begin{eqnarray}
\sigma^2_{b^{os}} \: =\: \frac{2 \: \sigma^2_{\eta^{os}_k}}{s^2}
\: =\: \frac{2 \: \alpha \sigma^2_{\eta_k^m}}{s^2} \: =\: \alpha \sigma^2_{b^m},
\label{eq:osvar}
\end{eqnarray}
where $\alpha$ is a reduction factor that depends on the order 
statistic and on the distribution of $b$. For most distributions, 
$\alpha$ can be found in tabulated form \cite{arba92}.
For example, Table~\ref{tab:alpha} 
provides $\alpha$ values for all order statistic combiners, up to $10$ 
classifiers, for a Gaussian distribution \cite{arba92,sagr56}. (Because this
distribution is symmetric, the $\alpha$ values of $l$ and $k$ where $l+k=N+1$
are identical, and listed in parenthesis).

Returning to the error calculation, we have: ${\cal M}_1^{os} = 0$, and 
${\cal M}_2^{os} = \sigma^2_{b^{os}}$, 
providing: 
\begin{eqnarray}
E_{model}^{os} = \frac{s {\cal M}_2^{os}}{2} = 
\frac{s \sigma^2_{b^{os}}}{2} = 
\frac{s \alpha \sigma^2_{b^m}}{2} = \alpha \; E_{model}^m.
\label{eq:red}
\end{eqnarray}

\begin{table} [htb] \centering
\caption{Reduction factors $\alpha$ for the Gaussian 
Distribution, based on \protect\cite{sagr56}.}
\vspace*{.1in} 
{\small 
\begin{tabular}{ccc|ccc|ccc} \hline
N & $k$ & $\alpha$ & N & $k$ & $\alpha$ & N & $k$ & $\alpha$ \\ \hline
1  & 1     & 1.00 & 6  & 2 (5) & .280 &    & 1 (9) & .357 \\ \cline{1-3}
2  & 1 (2) & .682 &    & 3 (4) & .246 &    & 2 (8) & .226 \\ \cline{1-6}
3  & 1 (3) & .560 &    & 1 (7) & .392 & 9  & 3 (7) & .186 \\ 
   & 2     & .449 & 7  & 2 (6) & .257 &    & 4 (6) & .171 \\ \cline{1-3}
4  & 1 (4) & .492 &    & 3 (5) & .220 &    & 5     & .166 \\ \cline{7-9}
   & 2 (3) & .360 &    & 4     & .210 &    & 1 (10)& .344 \\ \cline{1-6}
   & 1 (5) & .448 &    & 1 (8) & .373 &    & 2 (9) & .215 \\ 
5  & 2 (4) & .312 & 8  & 2 (7) & .239 & 10 & 3 (8) & .175 \\ 
   & 3     & .287 &    & 3 (6) & .201 &    & 4 (7) & .158 \\ \cline{1-3}
6  & 1 (6) & .416 &    & 4 (5) & .187 &    & 5 (6) & .151 \\ \hline
\end{tabular}
}
\label{tab:alpha}
\end{table}
Equation~\ref{eq:red} shows that the reduction in the error due to using
the OS combiner instead of the $m$th classifier is directly related to the
reduction in the variance of the boundary offset $b$. Since the means and
variances of order statistics for a variety of distributions are widely 
available in tabular form, the reductions can be readily quantified.

\subsection{Combining Biased Classifiers through Order Statistics}
\label{sec:osbi}
In this section, we analyze the error regions for biased classifiers. 
Let us return our attention to $b^{os}$. 
First, note that the error terms can no longer be studied separately,
since in general
$(a + b)^{os} \neq a^{os} + b^{os}$. We will therefore need to specify 
the mean and variance of the result of each operation\footnote{Since the
exact distribution parameters of $b^{os}$ are not known, 
we use the sample mean and the sample variance.}.
Equation~\ref{eq:bos} becomes:
\begin{eqnarray}
b^{os} \: = \: \frac{(\beta_i + \eta_i(x_b))^{os} - 
(\beta_j + \eta_j(x_b))^{os}}{s}.
\end{eqnarray}
Let $\bar{\beta_k} = \frac {1}{N} \;\sum_{m=1}^N \beta^m_k \:$ 
be the mean of classifier biases.
Since $\eta_k^m$'s have zero-mean, $\beta_k + \eta_k(x_b)$ has first 
moment $\bar{\beta_k}$ and variance
$\sigma^2_{\eta_k^m} + \sigma^2_{\beta_k^m}$, with 
$\sigma^2_{\beta_k^m} = E[(\beta^m_k)^2] - \bar{\beta_k}^2$, 
where $[\cdot]$ denotes the expected value operator.

Taking a specific order statistic of this
expression will modify both moments. The first moment is given
by $\bar{\beta_k} + \mu^{os}$, where $\mu^{os}$ is a shift which depends 
on the order statistic chosen, but not on the class. 
Then, the first moment of $b^{os}$ is given by:
\begin{eqnarray}
\frac{(\bar{\beta_i} + \mu^{os})  - (\bar{\beta_j} + \mu^{os})}{s} \: = 
\frac{\bar{\beta_i} - \bar{\beta_j}}{s} \: = \: \bar{\beta}.
\label{eq:bbias}
\end{eqnarray}
Note that the bias term represents an ``average bias'' since the
contributions due to the order statistic are removed. Therefore,
reductions in bias cannot be obtained 
from a table similar to Table~\ref{tab:alpha}.  

Now, let us turn our attention to the variance. 
Since $\beta_k^m + \eta_k^m(x_b)$ has variance 
$\sigma^2_{\eta_k^m} + \sigma^2_{\beta_k^m}$, it
follows that $(\beta_k + \eta_k(x_b))^{os}$ has variance 
$\sigma^2_{\eta_k^{os}} = \alpha (\sigma^2_{\eta_k^m} + \sigma^2_{\beta_k^m})$,
where $\alpha$ is the factor discussed in Section~\ref{sec:osno}. 
Therefore, the variance of $b^{os}$ is given by:
\begin{eqnarray}
\sigma^2_{b^{os}} & =& \frac{\sigma^2_{\eta^{os}_i} + \sigma^2_{\eta^{os}_j}}
{s^2}
\: =\: \frac{2 \: \alpha \sigma^2_{\eta_i^m}}{s^2} +  
 \frac{\alpha (\sigma^2_{\beta_i^m} +  \sigma^2_{\beta_j^m})}{s^2} \nonumber \\ 
& =& \alpha (\sigma^2_{b^m} + \sigma^2_{\beta^m}),
\label{eq:varos}
\end{eqnarray}
where $\sigma^2_{\beta^m} = \frac{\sigma^2_{\beta_i^m} + \sigma^2_{\beta_j^m}}
{s^2}$ is the variance introduced by the systematic errors 
of different classifiers.

We have now obtained the first and second moments of $b^{os}$,
and can compute the model error. Namely, we have 
${\cal M}_1^{os}\: = \: \bar{\beta}$ and 
$\sigma^2_{b^{os}} \: = \: {\cal M}_2^{os}- ({\cal M}_1^{os})^2$, leading to:
\begin{eqnarray}
E^{os}_{model}(\beta) & = & 
\frac{s}{2} \: {\cal M}_2^{os} \: = \: \frac{s}{2} \: 
(\sigma^2_{b^{os}} + \bar{\beta}^2) \\
& = & \frac{s}{2} \: 
(\alpha (\sigma^2_{b^m} + \sigma^2_{\beta^m})+ \bar{\beta}^2).
\label{eq:addbi}
\end{eqnarray}
The reduction in the error is more difficult to assess in this case. 
By writing the error as:
\begin{eqnarray*}
E_{model}^{os}(\beta) \: = \:  
\alpha \frac{s}{2} \: (\sigma^2_b + ({\beta^m})^2) \: + \:  
\frac{s}{2}  (\alpha \sigma^2_{\beta} + \bar{\beta}^2 - \alpha ({\beta^m})^2), 
\end{eqnarray*}
we get:
\begin{eqnarray}
E_{model}^{os}(\beta) \: = \: 
\alpha \: E_{model}^m(\beta) + \frac{s}{2} \: 
(\alpha \sigma^2_{\beta} + \bar{\beta}^2 - \alpha ({\beta^m})^2) .
\label{eq:addbi2}
\end{eqnarray}
Analyzing the error reduction in the general case requires knowledge
about the bias introduced by each classifier. Unlike regression problems
where the bias and variance contributions to the error are additive
and well-understood, in classification problems their interaction is
more complex~\cite{frie97}. Indeed it has been observed that ensemble 
methods do more than simply reduce the variance~\cite{scfr97}.

Based on these observations and Equation~\ref{eq:addbi2}, let us 
analyze extreme cases.
For example, if each classifier has the same bias,
$\sigma^2_{\beta}$ is reduced to zero and $\bar{\beta} = \beta^m$.
In this case the error reduction can be expressed as:
\begin{eqnarray*}
E_{model}^{os}(\beta) \: = \: 
\frac{s}{2} \: (\alpha \sigma^2_b + ({\beta^m})^2 \; = \;
\alpha E_{model}^m(\beta) + \frac{s(1 - \alpha)}{2} ({\beta^m})^2, 
\end{eqnarray*}
where $\alpha$ balances the two contributions to the
error. A small value for $\alpha$ will reduce the first component
of the error (mainly variance), while leaving the second term
untouched. The net effect will be 
very similar to results obtained for regression problems. In this
case, it is important to reduce classifier bias before
combining (e.g., by using an overparametrized model).

If on the other hand, the biases produce a zero mean variable,
we obtain $\bar{\beta} = 0$. In this case, the model 
error becomes:
\begin{eqnarray*}
E_{model}^{os}(\beta) \: = \: \alpha \: E_{model}^m(\beta) +
\frac{s \: \alpha}{2} \: (\sigma^2_{\beta^m} - ({\beta^m})^2)
\label{eq:erosbi}
\end{eqnarray*}
and the error reduction will be significant if the second term
is small or negative. In fact, if the variation among the biases
is small relative to their magnitude, the error will be reduced more
than in the unbiased cases. If however, the variation is large compared
to the magnitude, the error reduction will be minimal. Furthermore,
if $\alpha$ is large and the biases are small and highly varied, it
is possible for this combiner to do worse than the individual classifiers,
which is a danger not present for regression problems.
This observation very closely parallels results reported in~\cite{frie97}.

\section{Linear Combining of Ordered Classifier Outputs}
\label{sec:comb}
In the previous section, we derived error reductions when the 
class posteriors are directly estimated through the ordered
classifier outputs.
Since simple averaging has also been shown to provide benefits, in
this section, we investigate the  
combinations of averaging and order statistics for
pooling classifier outputs.

\subsection{Spread Combiner}
\label{sec:spread}
The first linear combination of ordered classifier outputs we study
focuses on extrema.
As discussed in
Section~\ref{sec:osno}, the maximum and minimum
of a set of classifier outputs carry specific meanings. Indeed, the maximum
can be viewed as the class for which there is the most evidence. Similarly, 
the minimum deletes classes with little evidence. In order to avoid a single
classifier from having too large of an impact on the eventual output,
these two values can be averaged to yield the {\em spread} combiner. This 
combiner strikes a balance between the positive and negative evidence,
leading to a more robust combiner than either of them.

\subsubsection{Spread Combiner for Unbiased Classifiers:}
\label{sec:spbi}
For a classifier without bias, the spread combiner is formally defined as:
\begin{eqnarray}
f_i^{spr}(x) = \frac{1}{2} \; (f_i^{1:N}(x) \; + \; f_i^{N:N}(x))
\; = \; p(c_i|x) \: + \:  {\eta}^{spr}_i(x) \:,
\label{eq:spr}
\end{eqnarray}
where:
\begin{eqnarray*}
{\eta}^{spr}_i(x) = \frac {1}{2} \; \left( 
\eta_i^{1:N}(x) \: + \: \eta_i^{N:N}(x) \right) \:.
\end{eqnarray*}
The variance of ${\eta}^{spr}_i(x)$ is given by:
\begin{eqnarray}
\sigma^2_{{\eta}^{spr}_i} & = & \frac{1}{4}  \sigma^2_{\eta^{1:N}_i(x)}
\; + \;  \frac{1}{4}  \; \sigma^2_{\eta^{N:N}_i(x)}
\; + \; \frac{1}{2} cov(\eta^{1:N}_i(x),\eta^{N:N}_i(x)).
\label{eq:etaspr}
\end{eqnarray}
where $cov(\cdot,\cdot)$ represents the covariance between two variables
(even when the $\eta_i$`s are independent, ordering introduces correlations).
Note that because of the ordering, the variances  in the first two terms
of Equation~\ref{eq:etaspr} can be expressed
in terms of the individual classifier variances.
Furthermore, the covariance between two order statistics can also be 
determined in tabulated form for given distributions. Table~\ref{tab:beta}
provides these values for a Gaussian distribution based on~\cite{sagr56}.
This expression can be further simplified for symmetric distributions
where $\sigma^2_{\eta^{1:N}} = \sigma^2_{\eta^{N:N}}$ (e.g., Gaussian noise
model) and leads to:
\begin{eqnarray}
\sigma^2_{{\eta}^{spr}_i} & = & 
\frac{1}{2}  \left( \alpha_{1:N} +  B_{1,N:N} \right) \sigma^2_{\eta_i(x)},
\label{eq:vars}
\end{eqnarray}
where $\alpha_{m:N}$ is the variance of the $m$th ordered sample
and $B_{m,l:N}$ is the covariance between the $m$th and $l$th
ordered samples,  given that the initial samples had unit 
variance~\cite{sagr56}. Because this is a symmetric distribution,
the $\beta$ values are also symmetric (e.g., $\beta_{1,2:5} = \beta_{4,5:5}$).

\begin{table} [htb] \centering
\caption{Some Reduction Factors $B$ for the Gaussian 
Distribution, based on \protect\cite{sagr56}.}
\vspace*{.1in}
{\small 
\begin{tabular}{ccc|ccc|ccc|ccc} \hline
N  & $k,l$   & $B$  & N  & $k,l$   & $B$  & N  & $k,l$   & $B$  & N
 & $k,l$   & $B$  \\ \hline
2 & 1,2 & .318 &   & 2,3 & .189 &   & 1,4 & .095 &   & 1,6 & .059 \\ \cline{1-3}
3 & 1,2 & .276 & 6 & 2,4 & .140 &   & 1,5 & .075 &   & 1,7 & .049 \\
  & 1,3 & .165 &   & 2,5 & .106 &   & 1,6 & .060 &   & 1,8 & .040 \\ \cline{1-3}
  & 1,2 & .246 &   & 3,4 & .183 &   & 1,7 & .048 &   & 1,9 & .031 \\ \cline{4-6}
4 & 1,3 & .158 &   & 1,2 & .196 &   & 1,8 & .037 &   & 2,3 & .154 \\
  & 1,4 & .105 &   & 1,3 & .132 &   & 2,3 & .163 &   & 2,4 & .117 \\
  & 2,3 & .236 &   & 1,4 & .099 & 8 & 2,4 & .123 &   & 2,5 & .093 \\ \cline{1-3}
  & 1,2 & .224 &   & 1,5 & .077 &   & 2,5 & .098 &   & 2,6 & .077 \\
  & 1,3 & .148 &   & 1,6 & .060 &   & 2,6 & .079 & 9 & 2,7 & .063 \\
5 & 1,4 & .106 & 7 & 1,7 & .045 &   & 2,7 & .063 &   & 2,8 & .052 \\
  & 1,5 & .074 &   & 2,3 & .175 &   & 3,4 & .152 &   & 3,4 & .142 \\
  & 2,3 & .208 &   & 2,4 & .131 &   & 3,5 & .121 &   & 3,5 & .114 \\
  & 2,4 & .150 &   & 2,5 & .102 &   & 3,6 & .098 &   & 3,6 & .093 \\ \cline{1-3}
  & 1,2 & .209 &   & 2,6 & .080 &   & 4,5 & .149 &   & 3,7 & .077 \\ \cline{7-9}
  & 1,3 & .139 &   & 3,4 & .166 &   & 1,2 & .178 &   & 4,5 & .137 \\
6 & 1,4 & .102 &   & 3,5 & .130 &   & 1,3 & .121 &   & 4,6 & .113 \\ \cline{4-6} \cline{10-12}
  & 1,5 & .077 &   & 1,2 & .186 & 9 & 1,4 & .091 &   &     &      \\
  & 1,6 & .056 & 8 & 1,3 & .126 &   & 1,5 & .073 &   &     &      \\ \hline
\end{tabular}
}
\label{tab:beta}
\end{table}

Then, using Equation~\ref{eq:sigb}, the variance of the boundary 
offset $b^{spr}$ can be calculated:
\begin{eqnarray}
\sigma^2_{b^{spr}} & = &  \nonumber
\frac {\sigma^2_{{\eta_i}^{spr}} \; + \; \sigma^2_{{\eta_j}^{spr}}} {s^2} \\ 
& = & 
\frac{1}{2}  \left( \alpha_{1:N} +  B_{1,N:N} \right) \sigma^2_{b}.
\end{eqnarray}
Finally, through Equation~\ref{eq:err}, we can obtain the reduction
in the model error due to the spread combiner:
\begin{eqnarray}
\frac{E_{model}^{spr}}{E_{model}}
\; = \; \frac{ \alpha_{1:N} +  B_{1,N:N} } {2} \; .
\label{eq:sprred}
\end{eqnarray}
Based on Equation~\ref{eq:sprred} and Tables~\ref{tab:alpha} and \ref{tab:beta},
Table~\ref{tab:redspr} displays the error reductions provided by the  spread 
combiner for a Gaussian noise model (for comparison purposes, the error reduction
for the $min$ and $max$ combiners is also provided. Note that for the Gaussian 
distribution, the error reduction of $min$ is equal to that of $max$.).

\begin{table} [htb] \centering
\caption{Error Reduction Factors for the Spread, $min$ and $max$ Combiners with Gaussian Noise Model.}
\vspace*{.1in} 
\begin{tabular}{c||c|c} \hline 
  N & $spread$  & $min$ or $max$   \\ \hline
 2 & .500 & .682 \\ 
 3 & .362 & .560 \\
 4 & .299 & .492 \\
 5 & .261 & .448 \\
 6 & .236 & .416 \\
 7 & .219 & .392 \\
 8 & .205 & .373 \\
 9 & .194 & .357 \\
10 & .186 & .344 \\ \hline 
\end{tabular}
\label{tab:redspr}
\end{table}

\subsubsection{Spread Combiner for Biased Classifiers:}
Now, if the classifier biases are non-zero, the spread combiner's
output is given by:
\begin{eqnarray}
f_i^{spr}(x) = \frac{1}{2} \; (f_i^{1:N}(x) \; + \; f_i^{N:N}(x))
\; = \; p(c_i|x) \: + \:  ({\eta}_i(x)  +  {\beta_i})^{spr}\:.
\label{eq:betaspr}
\end{eqnarray}
In that case, the boundary offset is given by:
\begin{eqnarray}
b^{spr} \: = \: \frac{(\beta_i + \eta_i(x_b))^{spr} -
(\beta_j + \eta_j(x_b))^{spr}}{s} \: ,
\end{eqnarray}
which after expanding each term and regrouping can be expressed as:
\begin{eqnarray}
b^{spr} & = &
\frac{ (\beta_i + \eta_i(x_b))^{1:N} - (\beta_j + 
\eta_j(x_b))^{1:N} }{2 s} \nonumber \\
& & + \; \;  \frac{
 (\beta_i + \eta_i(x_b))^{N:N} - (\beta_j + \eta_j(x_b))^{N:N} }{2 s} \: .
\label{eq:bbspr}
\end{eqnarray}

The first moment of $b^{spr}$ can be obtained by analyzing each term
of Equation~\ref{eq:bbspr}. In fact, the offset introduced by the first and
$n$th order statistic for classes $i$ and $j$ will cancel each other out,
leaving only the average bias between the min and max components
of the error (as in Equation~\ref{eq:bbias}), 
given by $\beta^{spr} = \frac{\beta_i^{1:N} - \beta_j^{1:N} + 
\beta_i^{N:N} - \beta_j^{N:N} }{s}$.

The variance of $b^{spr}$ needs to be derived from Equation~\ref{eq:bbspr}.
Proceeding as in Equation~\ref{eq:varos}, the variance of the spread
combiner can be expressed as:

\begin{eqnarray}
\sigma^2_{b^{spr}} & = & 
(\frac{1}{4}  \alpha_{1:N}  +  \frac{1}{4}  \alpha_{N:N} 
\; + \; \frac{1}{2} B_{1,N:N}) 
(\sigma^2_{b^m} + \sigma^2_{\beta^m}). 
\end{eqnarray}
For a symmetric distribution (where $\alpha_{1:N} = \alpha_{N:N}$), 
we obtain the following error:
\begin{eqnarray}
E_{model}^{spr}(\beta) & = & \frac{s}{2} {\cal M}_2 \;\; =\;\; 
\frac{s}{2} ( \sigma^2_{b^{spr}} \; + \; {{\cal M}_1}^2 ) \nonumber \\ 
& = & \frac{s}{2} \left
(\frac{1}{2}  \alpha_{1:N}  \; + \; \frac{1}{2} B_{1,N:N})
(\sigma^2_{b^m} + \sigma^2_{\beta^m}) + (\beta^{spr})^2 \right) \nonumber \\
& = & \frac{1}{2}  (\alpha_{1:N}  \; +  B_{1,N:N}) 
E_{model}(\beta) \: \: + \nonumber \\
& & \frac{s}{4} (\alpha_{1:N} + B_{1,N:N})  
		(\sigma^2_{\beta^m} - (\beta^m)^2)
+ \frac{s}{2}(\beta^{spr})^2 \; ,
\end{eqnarray}
which is very similar to Equation~\ref{eq:addbi2}, where the value
of $\alpha$ for a single order statistic is now replaced by
$\frac{\alpha_{1:N} + B_{1,N:N}}{2}$, since the mean of the first
and $n$th order statistic is used in the posterior estimate.

\subsection{Trimmed Means}
\label{sec:trim}
Instead of actively using the extreme values as was the case 
with the spread combiner, one can base the posterior estimate
around the median values. However, instead of selecting one
classifier output as was done for $f^{med}$, one can use multiple
classifiers whose outputs are ``typical.''
In this scheme, only a certain fraction of all available 
classifiers are used {\em for a given} pattern. The main advantage of 
this method over weighted averaging is that the set of classifiers
which contribute to the combiner vary from pattern to pattern. 
Furthermore, they do not need to be determined externally, but
are a function of the current pattern and the classifier responses
to that pattern.

\subsubsection{Trimmed Mean Combiner for Unbiased Classifiers:}
Let us formally  define the trimmed mean combiner 
($\beta_k = 0 , \forall k$) as follows:
\begin{eqnarray}
f_i^{trim}(x) = \frac{1}{N_2- N_1 +1} \; \sum_{m=N_1 }^{N_2} f_i^{m:N}(x)
\; = \; p(c_i|x) \: + \:  {\eta}^{trim}_i(x) \:,
\label{eq:trim}
\end{eqnarray}
where:
\begin{eqnarray*}
{\eta}^{trim}_i(x) = \frac {1}{N_2-  N_1 +1} \;\sum_{m= N_1 }^{N_2} \eta_i^m(x) \; .
\end{eqnarray*}
The variance of ${\eta}^{trim}_i(x)$ is given by:
\begin{eqnarray}
\! \! \! \! \! \! \! \! \! \! \sigma^2_{{\eta}^{trim}_i} 
\!\! & =  \!\! & \frac{1}{(N_2- N_1 +1)^2} \sum_{l= N_1 }^{N_2} 
\sum_{m= N_1 }^{N_2} cov(\eta^{m:N}_i(x),\eta^{l:N}_i(x)) \nonumber  \\
\!\! & =  \!\! &\frac{1}{(N_2 \! - \! N_1 \! +1)^2} \! \left( \sum_{m= N_1 }^{N_2} \!\! \sigma^2_{\eta^{m:N}_i(x)} \! \! + \! \! 
\sum_{m= N_1 }^{N_2} \sum_{l > m}^{N_2} 2 \; cov(\eta^{m:N}_i(x),\eta^{l:N}_i(x))
\right) \: .
\label{eq:ord}
\end{eqnarray}
Again, using the factors in Tables~\ref{tab:alpha} and \ref{tab:beta},
Equation~\ref{eq:ord} can be further simplified. 
Note that because the Gaussian distribution is symmetric, 
the covariance between the $k$th and $l$th ordered samples is 
the same as that between the $N+1-k$th and $N+1-l$th ordered samples.
Therefore, Equation~\ref{eq:ord} leads to:
\begin{eqnarray}
\sigma^2_{{\eta}^{trim}_i}  & = &  \frac{1}{(N_2- N_1 +1)^2} \sum_{m= N_1 }^{N_2} 
\alpha_{m:N} \; \; \sigma^2_{\eta_i(x)} \nonumber \\
& + & \frac{2}{(N_2- N_1 +1)^2} \sum_{m= N_1 }^{N_2} \sum_{l > m} 
 B_{m,l:N} \; \; \sigma^2_{\eta_i(x)} \: ,
\label{eq:ord2}
\end{eqnarray}
where $\alpha_{m:N}$ is the variance of the $m$th ordered sample
and $B_{m,l:N}$ is the covariance between the $m$th and $l$th
ordered samples,  given that the initial samples had unit 
variance~\cite{sagr56}.
Using the theory highlighted in Section~\ref{sec:error}, and
Equation~\ref{eq:ord2}, we obtain the following model error reduction:
\begin{eqnarray}
\frac{E_{model}^{trim}}{E_{model}} 
\; = \; \frac{1}{(N_2- N_1 +1)^2} 
\left( \sum_{m= N_1 }^{N_2} \alpha_{m:N}  \: + \:  
2 \:  \sum_{m= N_1 }^{N_2} \sum_{l > m} B_{m,l:N} \right) \; .
\label{eq:ordred}
\end{eqnarray}

Based on Equation~\ref{eq:ordred} and Tables~\ref{tab:alpha} and \ref{tab:beta},
we have generated a sample $trim$ combiner reduction table. Because there
are many possibilities for $N_1$ and $N_2$, a table that exhaustively provides
all reduction values is not practical.
In this sample table we have selected $N_1 = 2$ and $N_2 = N - 1$, that is, averaging
after the lowest and highest values have been removed. For comparison purposes
the reduction factors of the averaging combiner for $N$ and $N-2$ classifiers are
also provided (for i.i.d. classifiers the reduction factors are 1/N as derived
in~\cite{tugh96}; similar results were obtained for regression 
problems~\cite{peco93}). As these numbers demonstrate, although $N-2$ classifiers
are used in the trim combiner, {\em selectively} weeding out undesirable
classifiers provides reduction factors significantly better than simply averaging
$N-2$  arbitrary classifiers. The $trim$ combiner provides reduction factors
comparable the the $N$ classifier $ave$ combiner without being susceptible to
corruption by one particularly faulty classifier.

\begin{table} [htb] \centering
\caption{Error Reduction Factors for Trim and two corresponding $ave$ Combiners with Gaussian Noise Model.}
\vspace*{.1in} 
\begin{tabular}{c||c|c|c} \hline 
  N & $ave$ (for N) & $trim$  (for $N_1=2$ ; $N_2=N-1$) & $ave$ (for $N-2$) \\ \hline
 3 & .333 & .449 & 1.00 \\
 4 & .250 & .298 & .500 \\
 5 & .200 & .227 & .333 \\
 6 & .167 & .184 & .250 \\
 7 & .143 & .155 & .200 \\
 8 & .125 & .134 & .167 \\
 9 & .111 & .113 & .143 \\ \hline
\end{tabular}
\label{tab:redtri}
\end{table}

\subsubsection{Trimmed mean Combiner for Biased Classifiers:}
Now, if the classifier biases are non-zero, the trimmed mean combiner's
output is given by:
\begin{eqnarray}
f_i^{trim}(x) = 
\frac{1}{N_2- N_1 +1} \; \sum_{m=N_1 }^{N_2} f_i^{m:N}(x)
\; = \; p(c_i|x) \: + \: ({\eta}_i(x)  +  {\beta_i})^{trim}\:.
\label{eq:betatrim}
\end{eqnarray}
In that case the boundary offset is given by:
\begin{eqnarray}
b^{trim} \: = \: \frac{(\beta_i + \eta_i(x_b))^{trim} -
(\beta_j + \eta_j(x_b))^{trim}}{s}.
\end{eqnarray}

The first moment of $b^{trim}$ can be obtained from a manner similar
to that of the spread combiner.
Indeed,  each mean offset introduced by a specific order statistic 
for class $i$ will be offset by the one introduced for class $j$.
Only the trimmed mean of the biases will remain, giving the first 
moment of $b^{trim}$:
\begin{eqnarray}
\beta^{trim} = 
\frac{1} {N_2 - N_1 + 1} \sum_{m=N_1}^{N_2} 
\frac{\beta_i^{m:N} - \beta_j^{m:N}}{s} .
\end{eqnarray}

In deriving the variance of $b^{trim}$, we follow the same steps as in
Sections~\ref{sec:osbi} and \ref{sec:spbi}. The resulting boundary
variance is similar to Equation~\ref{eq:varos}, but the since the
reduction is due to the linear combination of multiple ordered
outputs, $\alpha$ is replaced by ${\cal A}$, where:
\begin{eqnarray}
{\cal A} = \frac{1}{(N_2- N_1 +1)^2}
\left( \sum_{m= N_1 }^{N_2} \alpha_{m:N}  \: + \:
		2 \:  \sum_{m= N_1 }^{N_2} \sum_{l > m} B_{m,l:N} \right) .
\label{aspr}
\end{eqnarray}

The model error reduction in this case is given by:
\begin{eqnarray}
E_{model}^{trim}(\beta) & = & \frac{s}{2} {\cal M}_2 \;\; =\;\; 
\frac{s}{2} ( \sigma^2_{b^{trim}} \; + \; {{\cal M}_1}^2 ) \nonumber \\ 
& = & \frac{s}{2} \left( {\cal A} \; \; 
   (\sigma^2_{b^m} + \sigma^2_{\beta^m}) + (\beta^{spr})^2 \right) \nonumber \\
& = & {\cal A} \; \; E_{model}(\beta) \: \: + \; \frac{s}{2} ({\cal A}  \; \;
		(\sigma^2_{\beta^m} - (\beta^m)^2) + (\beta^{spr})^2 )\; .
\end{eqnarray}
Once again we need to look at the interaction between the two parts 
of the error reduction. The first term
provides the error reduction compared to the model error 
of an individual classifier. The smaller ${\cal A}$ is, the more error
reduction there will be. In the second term, on the other hand, 
a small value for ${\cal A}$ is only useful if
the variability in the individual biases is higher than the biases
themselves ($\sigma^2_{\beta^m} > (\beta^m)^2$).  

\section{Experimental Results}
\label{sec:res}
The order statistics-based combining  methods proposed in this article 
are tailored for situations where:
\begin{enumerate}
\item{} individual classifier performance is uneven and class dependent;
\item{} it is not possible 
(insufficient data, high amount of noise)
to fine tune the individual classifiers without using computationally 
expensive methods.
\end{enumerate}
Such situations occur, for example,
in electrical logging while drilling for oil, where data from certain
well sites almost completely misses out on portions of the problem space,
and in imaging from airborne platforms where the 
classifiers receive inputs from different satellites and/or different types 
of sensors (e.g., thermal, optical, SAR).
While we have seen such data from Schlumberger, Austin, and NASA, Houston,
unfortunately the data sets are not standard or public domain.
So, in this article we restrict ourselves to public domain datasets and
simulate such variability by using ``early stopping'' i.e., prematurely
terminating the training of the individual classifiers\footnote{In all the
experiments reported here, ``high variability'' among classifiers refers
to classifiers being trained {\em exactly half as long} as the ``fine tuned''
classifiers.}.
Thus  combining results are first reported for the case
where only half the classifiers are finely tuned.
This procedure produces an artificially created {\em quality} variation 
in the pool of classifiers.

For the experiments reported below, we used a multi-layer perceptron (MLP)
with a single hidden layer, whose  weights were randomly initialized for
each run.  All classification results reported
in this article are {\em test set error rates averaged over 20
runs, along with the $95\%$ confidence intervals}.
Several types of 
simple combiners such as averaging, weighted averaging, voting,
median, products, weighted products (Bayesian),
using Dempster-Schafer theory of evidence, and entropy-based averaging, 
have been proposed in the literature.
However, on a wide variety of data sets, it has been observed that 
simple averaging usually provides results comparable to any of these techniques
(and, surprisingly,  often better than most of them)~\cite{ghtu92,tugh96b}.
For this reason, in this study, we use the average
combiner as a {\em representative} of simple combiners, for comparison
purposes.

The first two data sets (Tables~\ref{tab:sonvar} and \ref{tab:sonfin}) 
are based on underwater sonar signals. 
From the original sonar signals of four different underwater objects
(porpoise sound, cracking ice and two different whale sounds),
two feature sets are extracted~\cite{ghde92}:
\begin{itemize}
\item{} WOC: a 25-dimensional feature set, consisting of Gabor wavelet
coefficients, temporal descriptors and spectral measurements; and,
\item{} RDO: a 24-dimensional feature set, consisting of reflection 
coefficients based on both short and long time windows, and 
temporal descriptors.
\end{itemize}
For both feature sets, an MLP with 50 hidden units was used.
These data sets are available at URL {\em http://www.lans.ece.utexas.edu}.
Further details about this 4-class problem
can be found in~\cite{ghde92,tugh96b}. 

The next six data sets (Tables~\ref{tab:ucivar} and \ref{tab:ucifin}) 
were selected from the Proben1/UCI 
benchmarks~\cite{prec94}. The Proben1 benchmarks are particular
training, validation and test splits of the UCI data sets which are available 
from URL http://www.ics.uci.e\-du/\~{}m\-learn/ML\-Re\-posi\-tory.html. 
The results presented in this article are based on the first 
training, validation and test partition discussed in~\cite{prec94}, where
half the data is used for training, and a quarter each for validation and
testing purposes. 
Briefly these data sets, and the corresponding single layer feed-forward 
neural network architectures are\footnote{After deciding on a single
hidden layered architecture, the number of hidden units was determined
experimentally.}: \\
\begin{itemize}
\item{} Cancer: a 9-dimensional, 2-class data set
based on breast cancer data~\cite{mase90}, with 699 patterns; 
an MLP with 10 hidden units; 
\item{} Card: a 51-dimensional, 2-class
data set based on credit approval decision~\cite{quin87}, with
690 patterns; an MLP with 20 hidden units; 
\item{} Diabetes:
an 8-dimensional data set with two classes based on personal data 
from 768 Pima Indians obtained from 
the National institute of Diabetes
and Digestive and Kidney Diseases~\cite{smev88}; 
an MLP with 10 hidden units; 
\item{} Gene: a 120-dimensional data
set with two classes, based on the detection of splice junctions in
DNA sequences~\cite{noto91}, with 3175 patterns;
an MLP with 20 hidden units; 
\item{} Glass: a 9-dimensional, 6-class data set
based on the chemical analysis of glass splinters, with 214
patterns; an MLP with 15 hidden units; and, 
\item{} Soybean: an 82-dimensional, 19-class problem~\cite{mich80} 
with 683 patterns; an MLP with 40 hidden units.\\
\end{itemize}

\begin{table*} [htb] \centering
\caption{Combining Results in the Presence of High Variability in Individual 
Classifier Performance for the Sonar Data (\%~misclassified $\pm$ 95\% confidence interval).}
\vspace*{.1in} 
\begin{tabular}{c|c|c|c|c|c|c} \hline
Data   & N &  {Ave}      & {Max}  &  {Min} & {Spread} &  {Trim ($N_1$-$N_2$)} \\
 \hline \hline
RDO & 4 & 11.57 $\! \pm \!$ .22 & 11.94 $\! \pm \!$ .25 & 11.52 $\! \pm \!$ .40 & 11.04 $\! \pm \!$ .19 & 11.34 $\! \pm \!$ .28 (3-4) \\
13.32 $\! \pm \!$ 1.66 & 8 & 11.64 $\! \pm \!$ .18 & 11.47 $\! \pm \!$ .22 & 11.29 $\! \pm \!$ .27 & 11.51 $\! \pm \!$ .18 & 12.30 $\! \pm \!$ .17 (4-5) \\ \hline
WOC & 4 & 8.80 $\! \pm \!$ .18 & 7.84 $\! \pm \!$ .20 & 9.31 $\! \pm \!$ .24 & 8.54 $\! \pm \!$ .12 & 8.43 $\! \pm \!$ .26 (3-4) \\
12.07 $\! \pm \!$ 2.23 & 8 & 8.82 $\! \pm \!$ .17 & 7.68 $\! \pm \!$ .23 & 8.91 $\! \pm \!$ .13 & 8.24 $\! \pm \!$ .22 & 7.81 $\! \pm \!$ .16 (7-8) \\ \hline
\end{tabular}
\label{tab:sonvar}
\end{table*}

\begin{table*} [htb] \centering
\caption{Combining Results in the Presence of High Variability in Individual 
Classifier Performance for the Proben1/UCI Benchmarks (\%~misclassified $\pm$ 95\% confidence interval).}
\vspace*{.1in} 
\begin{tabular}{c|c|c|c|c|c|c} \hline
Data   & N &  {Ave}      & {Max}  &  {Min} & {Spread} &  {Trim ($N_1$-$N_2$)} \\
 \hline \hline
Cancer & 4 & 1.38 $\! \pm \!$ .13 & 1.38 $\! \pm \!$ .13 & 1.38 $\! \pm \!$ .13 & 1.38 $\! \pm \!$ .13 & 1.32 $\! \pm \!$ .13 (2-3) \\
1.49 $\! \pm \!$ .39 & 8 & 1.32 $\! \pm \!$ .12 & 1.44 $\! \pm \!$ .14 & 1.44 $\! \pm \!$ .14 & 1.44 $\! \pm \!$ .14 &  1.32 $\! \pm \!$ .12 (2-6) \\ \hline
Card & 4 & 13.60 $\! \pm \!$ .22 & 13.37 $\! \pm \!$ .22 & 13.49 $\! \pm \!$ .21 & 13.37 $\! \pm \!$ .22 & 13.60 $\! \pm \!$ .15 (3-4) \\
14.33 $\! \pm \!$ .36 & 8 & 13.66 $\! \pm \!$ .19 & 13.08 $\! \pm \!$ .14 & 13.02 $\! \pm \!$ .14 & 12.97 $\! \pm \!$ .12 & 13.20 $\! \pm \!$ .18 (7-8) \\ \hline
Diabetes & 4 & 25.26 $\! \pm \!$ .37 & 25.00 $\! \pm \!$ .46 & 25.00 $\! \pm \!$ .42 & 25.00 $\! \pm \!$ .42 & 25.26 $\! \pm \!$ .37 (3-4) \\
26.09 $\! \pm \!$ 1.27 & 8 & 24.84 $\! \pm \!$ .36 & 25.05 $\! \pm \!$ .33 & 25.05 $\! \pm \!$ .33 & 25.05 $\! \pm \!$ .33 & 24.84 $\! \pm \!$ .30 (6-8) \\ \hline
Gene & 4 & 12.90 $\! \pm \!$ .23 & 12.90 $\! \pm \!$ .26 & 12.94 $\! \pm \!$ .25 & 12.66 $\! \pm \!$ .21 & 12.67 $\! \pm \!$ .22 (3-4) \\
15.01 $\! \pm \!$ .78 & 8 & 12.89 $\! \pm \!$ .22 & 12.76 $\! \pm \!$ .24 & 12.41 $\! \pm \!$ .10 & 12.43 $\! \pm \!$ .22 & 12.56 $\! \pm \!$ .20 (7-8) \\ \hline
Glass & 4 & 33.77 $\! \pm \!$ .27 & 40.19 $\! \pm \!$ .72 & 33.21 $\! \pm \!$ .44 & 33.21 $\! \pm \!$ .44 & 33.77 $\! \pm \!$ .27 (2-3) \\
42.78 $\! \pm \!$ .75 & 8 & 33.96 $\! \pm \!$ .06 & 39.43 $\! \pm \!$ .27 & 33.77 $\! \pm \!$ .27 & 33.40 $\! \pm \!$ .41 &  33.77 $\! \pm \!$ .27 (1-6) \\ \hline
Soybean & 4 & 7.76 $\! \pm \!$ .11 & 7.94 $\! \pm \!$ .14 & 12.88 $\! \pm \!$ .39 & 7.71 $\! \pm \!$ .15 & 7.82 $\! \pm \!$ .18 (3-4) \\
10.71 $\! \pm \!$ 1.69 & 8 & 7.65 $\! \pm \!$ .00 & 7.82 $\! \pm \!$ .13 & 13.41 $\! \pm \!$ .53 & 7.71 $\! \pm \!$ .15 & 7.65 $\! \pm \!$ .00 (4-8) \\ \hline
\end{tabular}
\label{tab:ucivar}
\end{table*}

Tables~\ref{tab:sonvar} and \ref{tab:ucivar} present the combining results for
the Proben1 benchmarks and the underwater acoustic data sets
respectively, when the individual classifier performance was
highly variable. 
The misclassification percentage for individual classifiers are 
reported in the first column.
For the trimmed mean 
combiner, we also provide $N_1$ and $N_2$, the upper and lower cutting
points in the ordered average used in Equation~\ref{eq:trim}, obtained
through the validation set.

On the Sonar data, the results indicate that when the individual 
classifier performance is highly variable, order statistics-based 
combiners (particularly the {\em spread} combiner)
provide better classification results than
simple combiners. This performance improvement is obtained without
sacrificing the simplicity of the combiner. On the UCI/Proben1
benchmarks, the order statistics based combiners provide better
classification performance on three of the six sets studied
(no statistically significant differences were detected among the 
  various combiners in the remaining data sets).
One important thing to note, however, is that in all eight data sets
studied, the order statistics based combiners performed {\em at least
as well} as the simple combiner, implying that no risk is taken 
by using this method.

A close inspection of these results reveals that using either the
{\em max} or {\em min} combiner can provide better classification
rates than {\em ave}, but it is difficult to determine which of the 
two will be more successful given a data set. A validation set may be
used to select one over the other, but in that case, potentially 
precious training data is used solely for  determining which combiner
to use. The use of the {\em spread} combiner removes this dilemma by
consistently providing results that are comparable to, or better than,
the best of the {\em max-min} duo.
It is important to note that the {\em min} combiner performs poorly on the
Soybean data. Because this data set has 19 outputs, the posterior
estimates of unlikely classes become extremely small and highly
inaccurate. Basing decisions on such spurious values compromises
the combiner's performance. Notice, however, that the {\em spread} 
combiner is not adversely affected by this phenomenon.

\begin{table*} [htb] \centering
\caption{Combining Results with Fine-Tuned Classifiers 
for the Sonar Data (\%~misclassified $\pm$ 95\% confidence interval).}
\vspace*{.1in} 
\begin{tabular}{c|c|c|c|c|c|c} \hline
Data   & N &  {Ave}      & {Max}  &  {Min} & {Spread} &  {Trim ($N_1$-$N_2$)} \\ \hline \hline
RDO & 4 & 9.26 $\! \pm \!$ .32 &9.67 $\! \pm \!$ .20 &9.45 $\! \pm \!$ .19 &9.33 $\! \pm \!$ .20 &9.28 $\! \pm \!$ .28 (2-3) \\
9.95 $\! \pm \!$ .36     & 8 & 8.94 $\! \pm \!$ .06 &9.62 $\! \pm \!$ .16 &9.36 $\! \pm \!$ .15 &9.48 $\! \pm \!$ .18 & 8.92  $\! \pm \!$  .10 (1-6)\\ \hline
WOC & 4 & 7.05 $\! \pm \!$ .12 &7.31 $\! \pm \!$ .15 &7.44 $\! \pm \!$ .17 &7.31 $\! \pm \!$ .16 &7.05 $\! \pm \!$ .16 (2-3) \\
7.47 $\! \pm \!$ .21 & 8 & 7.17 $\! \pm \!$ .08 &7.19 $\! \pm \!$ .12 &7.41 $\! \pm \!$ .16 &7.22 $\! \pm \!$ .07 & 7.07  $\! \pm \!$ .10 (2-6)\\ \hline
\end{tabular}
\label{tab:sonfin}
\end{table*}

When there is ample data, and all the classifiers are finely tuned
(i.e., a validation set is used to determine the stopping time that 
  yields the best generalization performance),
simple combiners are expected to be adequate.
However, it
is not always possible to determine whether all conditions that
lead to such an ideal situation are satisfied. Therefore, it is
important to know whether the trimmed mean and spread combiners 
presented in this article perform worse than simple combiners
under such conditions. To that end, we have combined 
finely tuned feed forward neural networks using the methods proposed
in this article and compared the results with the traditional averaging method. 
In this new set of
experiments, all the conditions favor the averaging combiner (i.e., all
possible difficulties for the average combiner have been removed).
The results displayed in Tables~\ref{tab:sonfin} and \ref{tab:ucifin} 
indicate that, even under such circumstances,
both  the {\em spread} and {\em trim} combiners provide
results that are comparable to those obtained by the {\em ave} 
combiner. Furthermore, even under such conditions, the order statistics
combiners provide statistically significant improvements on two data sets.

\begin{table*} [htb] \centering
\caption{Combining Results with Fine-Tuned Classifiers 
for the Proben1/UCI Benchmarks (\%~misclassified $\pm$ 95\% confidence interval).}
\vspace*{.1in} 
\begin{tabular}{c|c|c|c|c|c|c} \hline
Data   & N &  {Ave}      & {Max}  &  {Min} & {Spread} &  {Trim ($N_1$-$N_2$)} \\ \hline \hline
Cancer & 4 & 0.69 $\! \pm \! $ .11 & 0.69 $\! \pm \!$ .11 & 0.69 $\! \pm \!$ .11 & 0.69 $\! \pm \!$ .11 & 0.69 $\! \pm \!$  .11 (2-3)\\
.69 $\! \pm \!$ .11 & 8 & 0.69 $\! \pm \!$ .11 & 0.57 $\! \pm \!$ .01 & 0.57 $\! \pm \!$ .01 & 0.57 $\! \pm \!$ .01 & 0.57 $\! \pm \!$ .11 (7-8)\\ \hline
Card & 4 & 13.14 $\! \pm \!$ .23 & 12.91 $\! \pm \!$ .11 & 13.02 $\! \pm \!$ .23 & 12.91 $\! \pm \!$ .11 & 13.14 $\! \pm \!$ .23 (2-3)\\
13.87 $\! \pm \!$ .36   & 8 & 13.14 $\! \pm \!$ .23 & 12.79 $\! \pm \!$ .01 & 12.79 $\! \pm \!$ .01 & 12.79 $\! \pm \!$ .01 & 12.80 $\! \pm \!$ .01 (7-8)\\ \hline
Diabetes & 4 & 23.33 $\! \pm \!$ .29 &23.23 $\! \pm \!$ .30 &23.33 $\! \pm \!$ .24 &23.23 $\! \pm \!$ .30 &23.33 $\! \pm \!$ .29 (3-4)\\
23.52 $\! \pm \!$ .35 & 8 & 22.92 $\! \pm \!$ .23 &23.23 $\! \pm \!$ .34 &23.12 $\! \pm \!$ .34 &23.23 $\! \pm \!$ .34 & 22.92  $\! \pm \!$  .23 (4-8)\\ \hline
Gene & 4 & 12.41 $\! \pm \!$ .21 &12.46 $\! \pm \!$ .24 &12.51 $\! \pm \!$ .18 &12.41 $\! \pm \!$ .17 &12.41 $\! \pm \!$ .12 (3-4)\\
13.49 $\! \pm \!$ .21 & 8 & 12.26 $\! \pm \!$ .14 &12.46 $\! \pm \!$ .18 &12.16 $\! \pm \!$ .08 &12.11 $\! \pm \!$ .19 &12.16 $\! \pm \!$  .09 (1-6)\\ \hline
Glass & 4 & 32.08 $\! \pm \!$ .01 &32.45 $\! \pm \!$ .36 &32.08 $\! \pm \!$ .01 &32.08 $\! \pm \!$ .01 &32.08 $\! \pm \!$ .01  (3-6)\\
32.26 $\! \pm \!$ .27  & 8 & 32.08 $\! \pm \!$ .01 &32.08 $\! \pm \!$ .01 &32.08 $\! \pm \!$ .01 &32.08 $\! \pm \!$ .01 & 32.08 $\! \pm \!$ .01 (3-6)\\ \hline
Soybean & 4 & 7.06 $\! \pm \!$ .00 &7.18 $\! \pm \!$ .11 &8.12 $\! \pm \!$ .77 &7.06 $\! \pm \!$ .00 &7.06 $\! \pm \!$ .00 (3-6)\\
7.36 $\! \pm \!$ .43     & 8 & 7.06 $\! \pm \!$ .00 &7.18 $\! \pm \!$ .05 &9.06 $\! \pm \!$ .82 &7.06 $\! \pm \!$ .00 & 7.06 $\! \pm \!$ .00 (3-6)\\ \hline
\end{tabular}
\label{tab:ucifin}
\end{table*}

\section{Conclusion}
\label{sec:disc}
In this article we present and analyze combiners based on
order statistics. These combiners 
blend the simplicity of averaging with the generality of
meta-learners. 
They are particularly effective if there are significant variations among
component classifiers in at least some parts of the joint
input-output space. Variations can arise when the individual
training sets cannot be considered as {\em random} samples from a
common universal data set. Examples of such cases include real-time
data acquisition and  classification
from geographically distributed sources or data mining problems with
large databases, where random subsampling is computationally expensive
and practical methods lead to non-random subsamples~\cite{brfa98}. 
Furthermore, The robustness of order statistics combiners is also helpful 
when certain individual classifiers experience catastrophic failures (e.g., 
due to faulty sensors).

The analytical framework provided in this paper quantifies
the reductions in error achieved when an order statistics 
based ensemble is used.  It also shows that 
the two methods for linear combination of order statistics introduced in this
paper provide more reliable estimates of the true posteriors than any of the
individual order statistic combiners.

The experimental results of Section 5 indicate that when
there is high variability among the classifiers, the order
statistics-based combiners significantly outperform simple
combiners, whereas in the absence of such variability these combiners
perform no worse. 
Thus the family of order statistic combiners
is able to extract an appropriate amount of information from the individual 
classifier outputs without requiring tuning additional parameters as in 
meta-learners, and without being substantially affected by outliers.

A future endeavor, which will be helpful for this work as
well as for the study of
classification based on very large datasets in general,
is to obtain a suite of public domain datasets which are intrinsically
partitioned into segments with varying quality. Though 
such situations sometimes  occur in practice (for example in oil logging
data \cite{chak97} and mortgage scoring \cite{merz98}; both data 
sets proprietary), they are not
represented in the standard, venerable databases such as UCI, ELENA
and Statlog typically used by the academic community.
Perhaps the recent CRoss-Industry Standard Process for Data Mining
(CRISP-DM) initiative will provide a satisfactory  solution to this
problem in the near future.

\nocite{diko95a,diko95b,madi97}
\nocite{leti93,rogo94,sokr96}
\nocite{rati97,tibs96a,wolp97}
\nocite{baxt92,xukr92,xujo95} 

\vspace*{.1in}
\noindent {\bf Acknowledgements:} 
This research was supported in part
by AFOSR contract F49620-93-1-0307, ARO contracts DAAH04-94-G0417 and
DAAH04-95-10494, and NSF grant ECS 9307632.

\end{document}